\documentclass[10pt,twocolumn,letterpaper]{article}

\usepackage{wacv}
\usepackage{times}
\usepackage{epsfig}
\usepackage{graphicx}
\usepackage{amsmath}
\usepackage{amssymb}
\usepackage{booktabs}
\usepackage{enumitem}

\usepackage{subcaption}
\usepackage{color}
\usepackage{multirow}
\usepackage{tabularx}
\usepackage{bm}
\usepackage{xcolor}
\usepackage[normalem]{ulem}

\usepackage{algorithm2e}
\RestyleAlgo{ruled}

\usepackage{sidecap}

\DeclareMathOperator*{\argmax}{arg\,max}

\newcommand{\myparagraph}[1]{\vspace{1.5pt}\noindent{\bf #1}}

\def\wacvPaperID{1395} % Enter the WACV Paper ID here

\wacvalgorithmstrack   % Uncomment this line if you are submitting to the Algorithms Track.
\wacvfinalcopy % *** Uncomment this line for the final submission

\ifwacvfinal
\usepackage[breaklinks=true,bookmarks=false]{hyperref}
\else
\usepackage[pagebackref=true,breaklinks=true,colorlinks,bookmarks=false]{hyperref}
\fi

\usepackage[capitalize]{cleveref}
\crefname{section}{Sec.}{Secs.}
\Crefname{section}{Section}{Sections}
\Crefname{table}{Table}{Tables}
\crefname{table}{Tab.}{Tabs.}
\begin{document}

%%%%%%%%% TITLE
\title{Urban Scene Semantic Segmentation with Low-Cost Coarse Annotation}

%\author{Anurag Das\inst{1}%\orcidID{0000-1111-2222-3333} 
	%\and
	%Yongqin Xian\inst{4*} \and
	%Yang He\inst{5}\thanks{The majority of the work was done when Yongqin Xian and Yang He were with MPI for Informatics.}%\orcidID{2222--3333-4444-5555}
	%\and Bernt Schiele \inst{1}
	%\and Zeynep Akata \inst{2,3}
	%}
	
	%\institute{$^1$MPI for Informatics, Saarland Informatics Campus \\ 
	%$^2$MPI for Intelligent Systems 
	%$^3$University of Tübingen \\ 
	%$^4$ETH Zurich $^5$Amazon
	%}

\author{Anurag Das$^{1,2}$, Yongqin Xian$^{5}$, Yang He$^6$, Zeynep Akata$^{3,4}$, Bernt Schiele$^{1,2}$ \\
\small $^1$MPI for Informatics, $^2$Saarland Informatics Campus, $^3$MPI for Intelligent Systems, $^4$University of T\"{u}bingen, $^5$ETH Z\"{u}rich, $^6$CISPA \\
\small \{andas,schiele\}@mpi-inf.mpg.de, yongqin.xian@vision.ee.ethz.ch, yang.he@cispa.saarland, zeynep.akata@uni-tuebingen.de 
}
%\and

\maketitle
\thispagestyle{empty}

%%%%%%%%% ABSTRACT
\begin{abstract}
   For best performance, today's semantic segmentation methods use large %variety of 
   and carefully labeled datasets, requiring expensive annotation budgets. In this work, we show that coarse annotation is a low-cost but highly effective alternative for training semantic segmentation models. Considering the urban scene segmentation scenario, we leverage cheap coarse annotations for real-world captured data, as well as %almost free full annotations for rendered 
   synthetic data to train our model and show competitive performance compared with finely annotated real-world data. % which is $~5\times$ more costly than ours.
   %aim to reduce this annotation effort by using available cheaper coarse annotations. 
Specifically, we propose a coarse-to-fine self-training framework that generates pseudo labels for unlabeled regions of the coarsely annotated data, using synthetic data to improve predictions around the boundaries between semantic classes, and using cross-domain data augmentation to increase diversity. 
    %and overall perfoamnce. 
%    
%    using a combination of synthetic data and coarse real data, that boosts the original coarse annotations by generating pseudo labels for the unlabeled regions of those images. Our self training framework efficiently uses the rendered data to provide the finer details to the coarse annotations. 
%We conduct extensive experiments on the challenging . 
Our extensive experimental results on Cityscapes and BDD100k datasets demonstrate that our method achieves a significantly better performance vs annotation cost tradeoff, yielding a comparable performance to fully annotated data with only a small fraction of the annotation budget. %Moreover, our framework can serve as a pretraining strategy to further im
Also, when used as pretraining, our framework performs better compared to the standard fully supervised setting.
   %In principle, we boost the original coarse annotations with the help of self-training, that we generate pseudo labels for the unlabeled regions of those images. 
   
   %In particular, the unlabeled regions usually have rich details and small objects. Therefore, we present a novel self-training framework that effectively makes fully use of synthetic data to provide important details and boundaries for generating better pseudo labels. We also make use of strong data augmentation that further improves our performance. \textbf{What is strong augmentation, please explain} 
   %We conduct experiments on widely used Cityscapes dataset and our approach achieve comparable performance to the model trained on all fine annotations. In addition to low-budget regime, our approach using coarse annotations only already outperforms using fine annotations only.
%   \bernt{updated abstract to also use the key terms boundary and augmentation}
%\keywords{Semantic Segmentation, self training, pseudo label, coarse annotation, data augmentation}
\end{abstract}

%%%%%%%%% BODY TEXT
\section{Introduction}
\label{sec:intro}

\begin{figure}[t]
\centering
\includegraphics[width=\linewidth]{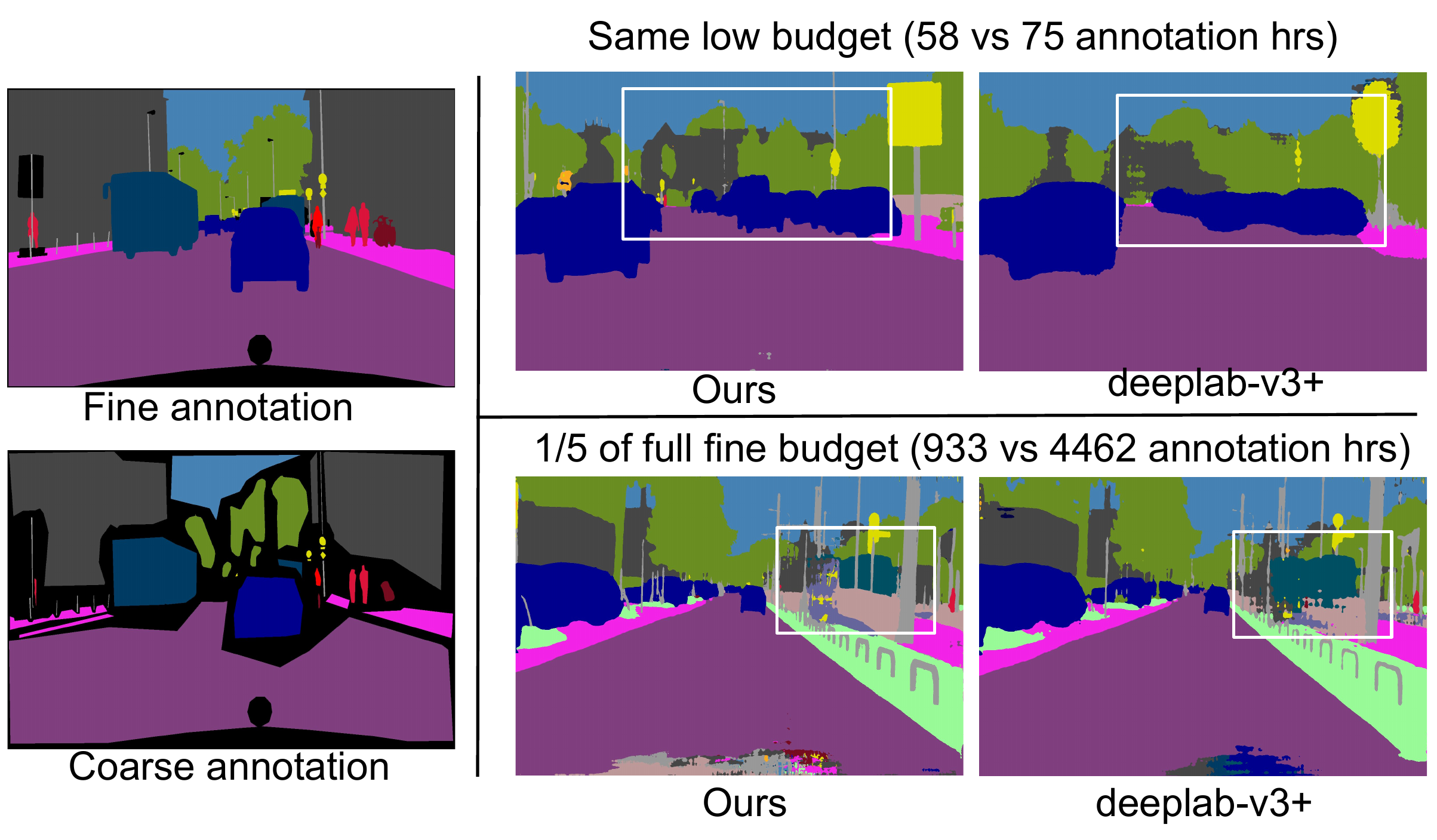}
\caption{%Coarse annotations are cheap but do not have fine details as compared to fine annotations. We propose a self training framework %\bernt{can we come up with a name for the approach?} 
%that improves the coarse labels, by generating pseudo labels for the ignored coarse label pixels.
%https://docs.google.com/drawings/d/1mZwGmzLfXmjvZtfKaxRF4pv8IvvceiIUwKJDMWSFQPw/edit?usp=sharing
%Left : In this example, in row 1 our framework can predict better than DeepLab-v3+ in the same low budget annotation regime (see car, building). Further in row 2, even with one fifth of full fine annotation budget, not only we perform comparable by overall performance, but also perform better for several tail distribution classes (e.g. train). Regions of interest are present in white bounding boxes. Right : Comparison between coarse and fine annotation. Coarse annotation trades off boundary accuracy for lower annotation cost. 
Left: Comparison between coarse and fine annotation. Coarse annotation trades off boundary accuracy for lower annotation cost. Right: Our method using coarse and synthetic data can predict better than DeepLab-v3+ trained on fine data in similar lower budget regime (see car and building in row 1), whereas with one-fifth of full fine budget our method perform comparable for most classes and even better for tail distribution classes (e.g. train in row 2). 
%In this example, in row 1 our framework can predict better than DeepLab-v3+ in the same low budget annotation regime (see car, building). Further in row 2, even with one fifth of full fine annotation budget, not only we perform comparable by overall performance, but also perform better for several tail distribution classes (e.g. train). Regions of interest are present in white bounding boxes.
}
\label{fig:teaser}
%\vspace{-4pt}
\end{figure}
Deep learning has made substantial progress in the field of semantic segmentation thanks to the availability of large-scale datasets~\cite{cordts2016cityscapes,everingham2010pascal,lin2014microsoft}. However, annotating those datasets for semantic segmentation requires carefully labeling every pixel in the images, which is time-consuming and expensive. This has motivated abundant attempts to reduce the annotation efforts by exploring weaker forms of supervision, for example image-level label~\cite{oh2017exploiting,huang2018weakly}, bounding box~\cite{khoreva2017simple}, scribble~\cite{lin2016scribblesup} and points~\cite{bearman2016s}. Those works mainly focus on the PASCAL VOC dataset~\cite{everingham2010pascal} with only a few object instances per image, which is a relatively easy scenario. In real-world urban scenes, however, the density of traffic participants is significantly higher~\cite{cordts2016cityscapes}. Reducing the annotation burden for such complex urban scenarios remains  challenging and underexplored. Cityscapes~\cite{cordts2016cityscapes} is one of the most popular datasets for urban scene segmentation, consisting of two kinds of pixel-level annotations (see \cref{fig:teaser}): 
(1) Coarse annotation - annotators draw coarse polygons for labelling classes ignoring the finer details around class boundaries. The goal is to annotate as many pixels as possible within 7 min of annotation time per image~\cite{cordts2016cityscapes}, with the only condition that each polygon must have pixel labels from a single class.  
(2) Fine annotation - annotators draw fine polygons that align well with the object boundaries, which is much more time-consuming i.e., 90 minutes per image. Coarse annotations tradeoff finer details for lower annotation cost.   
%was obtained by labeling layered polygons~\cite{russell2008labelme} with precise boundaries, which takes more than 90 minutes for a single image, and (2) coarse annotation was implemented by labeling coarse polygons with unlabeled regions~(see \cref{fig:teaser}), which takes only 7 minutes per image. 
For best performance, most existing works~\cite{fcn2015cvpr,chen2018encoder,zhao2017pyramid,yuan2021segmentation,zhang2021dcnas,yin2020disentangled,zhao2018psanet,yu2020context,hou2020strip,takikawa2019gated,zhang2019acfnet} rely on the full fine annotations, which is significantly expensive to annotate. The coarse data is sometimes used as additional training data to marginally boost the performance~(less than $2\%$)~\cite{chen2018encoder,yuan2021segmentation,yin2020disentangled}. 
%While considerably advances were achieved with fine annotation~\cite{fcn2015cvpr,chen2018encoder,zhao2017pyramid,yuan2021segmentation, zhang2021dcnas, yin2020disentangled, zhao2018psanet,yu2020context,hou2020strip,takikawa2019gated,zhang2019acfnet}. 
%While results are remarkable, the annotation cost becomes tremendous. 
%coarse annotation is largely unexplored in urban scene segmentation. Prior works either ignore coarse data~\cite{zhao2017pyramid} or treat it equally with the fine data as additional training data~\cite{chen2018encoder}.  
However, relatively little is known about the potential value of using coarse annotations. %In particular, a natural question is whether 
In this work, we propose a novel hybrid supervision scheme to combine coarse data and synthetic data, aiming to reduce the annotation cost for urban scene segmentation without sacrificing final performance. 
%To this end, we propose a novel hybrid supervision scheme to combine coarse data and synthetic data. 
%We study the problem of urban scene segmentation using coarse data as supervision.
While coarse data has the advantages of being significantly larger~(over $10\times$) than fine data given a fixed annotation budget, it is notoriously difficult to achieve a good performance using coarse data alone, which partially explains the lacking literature in this area. One limitation is that the coarse data has a lot of unlabeled regions which might not contain sufficient supervision signals. We propose a coarse-to-fine self-training framework that generates pseudo labels with consistency constraints for unlabeled regions, gradually converting sparse coarse annotation into dense fine annotation. To circumvent the need of boundary information, we propose to leverage synthetic data which provides precise dense annotation and is free in terms of annotation cost. A boundary loss is applied on the synthetic data to encourage the network
to focus on fixing boundary errors. To alleviate the domain gap, we further perform the cross-domain data augmentation that mixes the images from two domains. %generate pseudo labels for unlabeled regions in coarse data following consistency constraints. 
Finally, we retrain our network with pseudo labelled coarse data and synthetic data to refine the network.

Our work makes the following contributions. First, we emphasize the potential value of coarse annotation, which is significantly cheaper than fine annotation, but has been largely ignored as a primary source for training. Second, we develop a strong baseline for urban scene segmentation %by leveraging three strong techniques, namely self-training, boundary loss and cross-domain data augmentation. This baseline 
that uses hybrid supervision signals from coarse and synthetic data therefore substantially reducing the annotation cost. To the best of our knowledge, we are the first to combine coarse and synthetic data for urban scene segmentation. Finally, we show the trade-off between the annotation budget and performance on the challenging Cityscapes and BDD100k datasets. We empirically demonstrate that our method consistently outperforms its fine data counterpart when using comparable annotation budgets. Notably, we achieve competitive performance with only one-fifth and one-eighth annotation cost compared to using fine annotation on Cityscapes and BDD100k datasets respectively.   

\section{Related Work}
\label{sec:rel_work}

\myparagraph{Semantic segmentation with weak annotations.}
%Deep learning based semantic segmentation \cite{fcn2015cvpr,chen2018encoder} requires large-scale training data to achieve decent performance.
%However, data annotation for semantic segmentation is rather expensive, demanding for detailed dense annotation. 
%A lot of efforts have been made to utilize weak annotations and reduce annotation efforts.
Prior works use image-level annotation~\cite{em_wss2015weakly,oh2017exploiting}, point~\cite{bearman2016s}, scribble~\cite{lin2016scribblesup} and bounding box~\cite{khoreva2017simple} annotations. Further, other works incorporate zero-shot learning~\cite{bucher2019zero,xian2019semantic,li2020consistent,das2021net} to transfer knowledge from categories with annotations to the novel classes during test.
However, prior works in this direction does not perform complex scene segmentation and only show results in segmenting few objects from images. %, while complex scene segmentation remains challenging. In urban scenes, many images have very similar label distributions. For instance, almost all the images contain the road class, therefore, image-level annotation is not as  powerful as object-level segmentation in this scenario. 
Different from these works, we focus on challenging urban scene semantic segmentation with coarse annotation.%, which needs to predict semantic class labels for every pixel. 
We show a feasible solution to reach competitive results with coarse annotations, which only needs 7 minutes for each 1024$\times$2048 image in average.

\myparagraph{Semi-supervised semantic segmentation.}
 studied for example in~\cite{hung2018adversarial,souly2017semi,chen2021semi,li2021semantic,chen2020naive,he2017std2p}, aims to leverage unlabeled data to improve the segmentation performance or reduce the annotation efforts. Generative adversarial training has been exploited to achieve this goal, either by applying a trained discriminator to provide training signals for unlabeled images~\cite{hung2018adversarial} or generate labeled pairs from GANs~\cite{li2021semantic}. In addition, other works propagate labels from a learned model to unlabeled images~\cite{chen2020naive,zhu2019improving,he2017std2p}.
 Even though the above approaches show success in reducing annotation efforts, they still need fine annotations to train a semantic segmentation model for urban scenes, while we only apply coarse annotation to obtain high-quality segmentation.

\myparagraph{Self-training and pseudo labels}
Self-training aims to learn from unlabeled data and generate pseudo labels
%on it to turn unsupervised learning to 
for supervised learning. %A popular method is to use a student-teacher approach.
%the model trained on labeled data as teacher model to generate pseudo labels and then train the student model with these pseudo labels. 
Xie~\etal~\cite{Xie_2020_CVPR_Noisy_student} present Noisy Student Training, in which the student model is added with noise such as dropout, stochastic depth and data augmentation. Ghiasi~\etal~\cite{Ghiasi_2021_ICCV_multi} introduce multi-task self-training (MuST) which uses several specialized teacher models trained on labeled data to create a multi-task pseudo labeled dataset. The dataset is used to train a single student model with multi-task learning. Zoph~\etal~\cite{nips2020_Rethinking} reveal that self-training is always helpful when using stronger data augmentation and in the case that pre-training is helpful, self-training improves upon pre-training. %Self-training is important for many computer vision tasks.
More generally, self-training has been exploited to improve the performance of semantic segmentation~\cite{chen2020naive} significantly.
%. By applying unlabeled images, segmentation performance is boosted a lot. 
Different to this work, we do not touch the fine annotated data to train our model and still obtain competitive performance at much lower annotation costs.

\begin{figure*}[t]
\centering
\includegraphics[width=\linewidth]{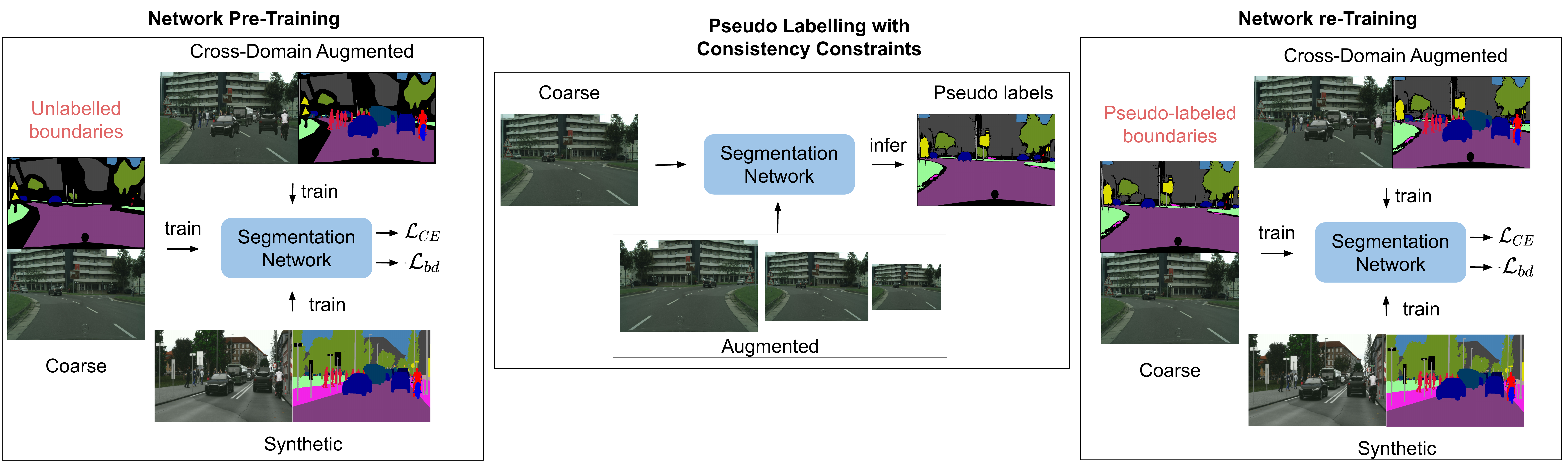}
\caption{Our coarse-to-fine self-training framework. We propose to improve the coarse annotations by generating pseudo labels. In the Network Pre-Training phase, we use coarse data with unlabelled boundaries and augment synthetic data. We have $\mathcal{L}_{CE}$, classification loss for all data, with additional $\mathcal{L}_{bd}$, boundary loss only for fine detailed synthetic data. After Network Pre-Training, we generate pseudo labels for ignored boundary regions in coarse annotations, followed by Network Re-Training, where we replace the coarse annotation with improved coarse annotations with pseudo labeled boundaries and train iteratively. 
}
%At test time, we generate pseudo labels by following consistency rules. During next iteration of training we replace the coarse labels with generated pseudo labels. (CE : cross entropy loss). 
%https://docs.google.com/drawings/d/1A7FZDGnk7jKtym5q4GH-jo0uxlBtMsQpzs4Wc7tmHNc/edit?usp=sharing

\label{fig:main_fig}
%\vspace{-4pt}
\end{figure*}

\myparagraph{Semantic segmentation using synthetic data}
%Synthetic data from game engines is much easier to obtain than carefully labeled real-world data. Once a rendering system is developed, it is almost free to obtain numerous high-quality data with full annotations. 
There are many successful datasets released for urban scenes~\cite{richter2016playing,wrenninge2018synscapes,ros2016synthia}. However, most prior works focus on pretraining a model~\cite{richter2016playing,wrenninge2018synscapes}, domain adaptation~\cite{tranheden2021dacs,tsai2018learning,chang2019all} or generalization~\cite{choi2021robustnet,yue2019domain}.
In this work, we combine real-world data with coarse annotations and synthetic data to train a semantic segmentation model for urban scenes. We make use of synthetic data to provide useful details and boundaries to networks, to predict pseudo labels for the unlabeled regions of coarsely annotated data. We show that just using coarse annotations along with synthetic data can perform comparable with a fraction of annotation budget of expensive fine annotations.

\section{Coarse-to-Fine Self-Training Framework}
Although coarse annotations could significantly reduce the labelling cost, it is notoriously difficult to achieve a good performance for urban scene segmentation using solely coarse data due to the missing boundary information. To this end, we propose a novel coarse-to-fine self-training framework that utilizes hybrid supervision from coarse and synthetic data and generates pseudo labels at unlabelled regions for re-training the network.  \cref{fig:main_fig} shows an overview of our coarse-to-fine self-training framework, consisting of three stages: 1) network pre-training using sparse coarse annotation and synthetic data, 2) pseudo labelling unlabeled regions in coarse data with consistency constraints, 3) network re-training with pseudo-labelled coarse data and synthetic data. The last two stages can also be performed iteratively to refine the network.

\subsection{Segmentation network pre-training}
The first stage of our framework is segmentation network pre-training where we learn a strong ``teacher'' network, e.g., DeepLab-v3+~\cite{chen2018encoder}, for pseudo labelling in the second stage. Let $D=(X_c, Y_c)\cup (X_s, Y_s)$ be our training set where $X_c$ denotes real images with coarse annotations $Y_c$, and $X_s$ denotes synthetic images with fine annotations $Y_s$. While coarse annotations $Y_c$ are particularly sparse at object boundaries, synthetic annotations $Y_s$ provide dense labels for every pixel in the synthetic images. The key insight of our approach is to capture real distributions from coarse data and learn boundary information from synthetic data. In the following, we describe our cross-domain data augmentation for reducing the domain gap, and the boundary loss for explicit boundary modeling.

\myparagraph{Cross-domain augmentation.} \label{pg:cda} Inspired by DACS~\cite{tranheden2021dacs}, we randomly sample class masks (with probability, $p=0.5$) from synthetic labels and corresponding image segments and paste them onto real masks and images (see Fig.2 and sec.1 of supplement). We select real samples with probability, $p=0.5$ from a given training batch and perform this augmentation using the synthetic data sampled uniformly from the whole synthetic dataset, to obtain a new set of training data $(X_{aug}, Y_{aug})$, i.e. $(X_{aug}, Y_{aug}) = (mask)\otimes(X_c,Y_c) + (1-mask)\otimes(X_s,Y_s)$
%\begin{align}
%\begin{split}
%    (X_{aug}, Y_{aug}) = (mask)\otimes(X_c,Y_c) \\ + (1-mask)\otimes(X_s,Y_s)
%\end{split}
%\end{align}
where, $mask$ is sampled from synthetic label and $\otimes$ is elementwise multiplication. This simple strategy alleviates the domain gap and improves data diversity, which is particularly helpful in low-budget regimes.
~\cite{tranheden2021dacs} solves unsupervised domain adaptation problem, where labels are not available for real images, whereas for our problem we have manually annotated coarse labels for the real images.
%Different from~\cite{tranheden2021dacs}, we paste synthetic class masks on real coarse labels(+pseudo labels in retraining phase), whereas ~\cite{tranheden2021dacs} paste synthetic class masks on generated pseudo labels. 
Also, we sample synthetic data for augmentation from whole dataset whereas~\cite{tranheden2021dacs} sample synthetic data only from the training batch.  

\myparagraph{Classification loss.} We adopt the standard cross-entropy loss~($\mathcal{L}_{CE}$) on the mixture of coarse data, synthetic data and augmented data, i.e., $(X_c, Y_c) \cup (X_s, Y_s) \cup (X_{aug}, Y_{aug})$. 

\myparagraph{Boundary loss.} As coarse annotations lack proper boundaries,  we propose to adopt a boundary loss on the synthetic data, encouraging the network to predict better boundaries. Suppose $\hat{y}_s$ is the predicted label mask of a synthetic image $x_s$ with GT mask $y_s$, we compute the prediction boundary $\Gamma_{pred}$ and the ground truth boundary $\Gamma_{GT}$ using the following equation, 
\begin{align}
    \Gamma_{pred} = ||\nabla \hat{y}_s||_2 \text{; } \Gamma_{GT} = ||\nabla y_s||_2  
\end{align}
where $\nabla$ is the gradient operator with central difference approximation. We estimate $\hat{y}_s$ with Gumbel Softmax trick~\cite{jang2016categorical} to make it differentiable for backpropagation. 
For both boundaries, we select pixels with representative boundaries by thresholding them (threshold=$1e^{-8}$~\cite{takikawa2019gated}). %We use the same threshold parameter as \cite{takikawa2019gated}. 
Assuming $p^+_{GT}$ and $p^+_{pred}$ as the corresponding boundary pixels for ground truth and segmentation prediction masks after thresholding (ie. $p^+_{GT}, p^+_{pred} \text{ are points where } \Gamma_{GT}>1e^{-8},  \Gamma_{pred}>1e^{-8}$ respectively), we calculate our boundary loss as:
\begin{align}
\begin{split}
    \mathcal{L}_{bd} = \lambda_1 |\Gamma_{pred} (p^+_{GT}) - \Gamma_{GT} (p^+_{GT}) | \\ + \lambda_2 |\Gamma_{pred} (p^+_{pred}) - \Gamma_{GT} (p^+_{pred}) |
\end{split}
\end{align}
where we set $\lambda_1, \lambda_2$ to be 0.5 in our experiments. This boundary loss has been used in~\cite{takikawa2019gated} by applying it on finely annotated images. In contrast, we apply this boundary loss on synthetic data and we do not access any finely annotated data, which is a more challenging setting. This loss term complements the cross-entropy loss by enforcing boundary consistency. We do not apply the boundary loss on coarse data due to its inaccurate boundaries.  %Further from results we see this boundary loss successfully transfers boundary information from synthetic data to coarse data with no boundaries by improving the performance along boundaries (supplement Sec.6 and \cref{fig:ablation1}~(right)).

\subsection{Iterative pseudo labelling and network re-training.}
%Our framework consists of three iterative steps. In step 1, first we perform  cross domain augmentation to help reduce the domain gap between the synthetic and real data. This also helps in improving the diversity in data, and is particularly helpful for lower budget regimes. After augmentation, all three data-types - coarse, synthetic and augmented data are fed to a segmentation network. We optimise the standard cross entropy loss~($\mathcal{L}_{CE}$) for all labeled pixels. %, i.e. $\mathcal{L}_{CE}(x_i, y_i) = -logP_{y_i}(x_i)$. 
%For obtaining important boundary information, we propose to use an additional boundary loss $\mathcal{L}_{bd}$, which is only applied on synthetic data as it has proper boundaries. As the name suggests, we apply this loss only on the class boundaries in the images. We discuss both cross-domain augmentation and boundary loss in detail in the following subsections. 
In the second stage, we use the model trained in the previous stage to generate pseudo labels for unlabelled pixels for the coarse training data. We adopt test time augmentation consistency~\cite{chen2020naive,pastore2021closer} for generating precise pseudo labels. Specifically, we use a combination of flip (flip, no flip) and resize (scale : 0.5, 1, 2.0) for augmentations. For any of the 6 combinations, if pseudo labels disagree, we mark the pixel as ignore. Further we also do confidence thresholding. If prediction confidence obtained after averaging the logits for 6 augmentations, is greater than the threshold (0.9), we accept the pseudo label, otherwise it is marked ignore.  

After generating the pseudo labels, we move to the third stage, where we replace the unlabelled pixels in the coarse-annotated images with the pseudo labels. Note that after each iteration we only replace the previous labels from the ignore regions of the coarse data and keep the manually annotated coarse labels intact. Then we re-train the segmentation network using the new coarse data, original synthetic data and augmented data with the same loss functions defined in the first stage. These two stages can be repeated in an iterative manner to refine the network. %After step 3, we again repeat all three steps in an iterative manner.  

%Using only coarse data to generate pseudo label doesn't provide better labels around the boundaries. It becomes difficult to achieve good performance using coarse data alone. So, we use synthetic data to provide us with the useful boundary information. Also, using synthetic data is not trivial. There is a domain gap between both data-types. We propose cross domain augmentation to help reduce the domain gap.
%We do cross-domain augmentation as presented in \ref{pg:cda} on coarse annotation. For each training batch, we have three different data types ie %\bernt{really `eg'? not rather `i.e.'?}
%- Coarse data, Synthetic data and Augmented data. We optimise for cross entropy loss $\mathcal{L}_{CE}(x_i, y_i) = -logP_{y_i}(x_i)$ for all images and apply boundary loss $\mathcal{L}_{bd}$ on images from synthetic dataset. Once training is done, we use the trained model to generate pseudo labels for the ignore regions. We adapt test time augmentation consistency as described in \ref{pg:plc} for generating correct pseudo labels. We replace coarse data with generated pseudo labelled data and repeat the iterative process as presented in \ref{alg:self_tr}.
\begin{algorithm}
\footnotesize
\caption{coarse-to-fine framework for generating pseudo labels}\label{alg:self_tr}
\KwData{ Coarse data : $(X_c, Y_c)$, Synthetic data : $(X_s, Y_s)$}
\textbf{Step 1 - Network Pre-Training} : \\
\begin{itemize}
    \item Do cross domain augmentation on coarse data to get $(X_{aug}, Y_{aug})$. 
    \item Train segmentation model $f_{\theta}(x)$ on combined data $(X, Y) = (X_c, Y_c) \cup (X_s, Y_s) \cup (X_{aug}, Y_{aug}) $
    \begin{align*}
\theta^* = \argmax_{\theta} \mathcal{L} (Y, f_{\theta}(X))
\end{align*}
where, $\mathcal{L} = \mathcal{L}_{CE} (Y, f(X, \theta)) + \lambda \mathcal{L}_{bd} (Y_s, f(X_{s}, \theta))$
\end{itemize}
\textbf{Step 2 - Pseudo Labelling with Consistency Constraints} : Generate pseudo labels from trained network $f_{\theta^*}$ for ignored region $(X_{ps}, Y_{ps})$, following consistency rules. \\
\textbf{Step 3 - Network Re-Training} : Replace coarse data $(X_c, Y_c)$ with pseudo labels $(X_{ps}, Y_{ps})$ and retrain iterating steps 2-3

\end{algorithm}

%\subsection{Strong Augmentation}
%\label{subsec:sa}

%\begin{figure}[t]
%\centering
%\includegraphics[width=0.5\linewidth]{aug.pdf}
%\caption{Left : DACS based cross-domain augmentation. In this example, we cut-paste car and bus from synthetic data to Coarse data. The augmented class objects are marked with white bounding boxes. Right : We perform morphological erosion on fine annotation followed by polygon fitting to mimic coarse annotation. Erosion removes rich boundary information from fine annotation. 
%}
%\vspace{-4pt}
%\label{fig:aug}
%\end{figure}

\subsection{Tackling class imbalance with model-based sampling}
\label{subsec:select}
%\bernt{as mentioned before the term `balanced sampling' is not particularly intriguing- can we change this?}
Selection of training samples is not trivial. Specifically for lower annotation budgets, where tail distribution classes appear scarcely, it becomes more important to have such samples in training. Also, since we assume that initially, images are unlabelled, it becomes difficult to get training samples from tail end classes. There is an additional manual classification overhead cost for obtaining the class distribution in the training data. We present model-based sampling to ensure pixels from tail end classes are sufficiently present in the training dataset without any manual overhead cost.

Our model-based sampling method makes use of a model trained on an initial set of randomly sampled, 1,000 coarsely annotate images. Using this initial model we estimate the class distribution of available unlabeled images. With the estimated class distribution information, we make sure that we have sufficient samples from each class by sampling almost the same number of data samples having a particular class. We build incrementally on the training samples, i.e., for 2,000 training samples, we use the initial 1,000 training samples and add another 1,000 samples obtained with the help of the initial segmentation model. We compare the performance of our sampling technique with random sampling in \cref{tab:iteration} and observe that our model-based sampling can indeed increase performance significantly.

{
 \setlength{\tabcolsep}{3pt}
 \renewcommand{\arraystretch}{1.1}
 \begin{table}
  \centering
  \resizebox{0.9\linewidth}{!}{%
\begin{tabular}{l|c|c|c | c | c }
Dataset & Type & train & val  & time & \% annotated  \\ 
\hline
\multirow{2}{*}{Cityscapes}  & Coarse & 19998 & NA & 7 min & 63.04 \\
& Fine & 2975 & 500 & 90 min & 99.98\\
\hline
\multirow{2}{*}{BDD100k}  & Coarse & 4000 & NA & 7 min  & 69.81 \\
& Fine & 3000 & 1000 & 75 min & 100\\
\hline 
Synscapes & Synthetic & 25000 & NA & NA & 100\\
\hline 
GTA-5 & Synthetic & 24966 & NA &  NA & 100\\
\end{tabular}
}
\caption{Left: Datasets statistics. train: number of training images, val: number of validation images, time: time to annotate an image, ``\% annotated'': average percentage of pixels annotated per image. Coarse annotation with less labeled pixels are significantly faster to annotate compared to fine annotations. }
\label{tab:stat}
\end{table}
}
%	\end{minipage}\hfill
%	\quad
%	\begin{minipage}{0.47\linewidth}
%		\centering
%		%\captionof{figure}{Caption_2}
%		\includegraphics[width=\textwidth]{sim1.pdf}
%	\end{minipage}
%	\caption{Left: Datasets statistics. train: number of training images, val: number of validation images, time: time to annotate an image, ``\% annotated'': average percentage of pixels annotated per image. Coarse annotation with less labeled pixels are significantly faster to annotate compared to fine annotations. 
%	}
%	\label{fig:sim}
%\end{figure}
%}
\section{Experiments}
\label{sec:exp}
In this section, we first describe our experimental setting, then we present our results and comparison with baselines, ablation studies with framework components followed by a qualitative analysis.

\myparagraph{Datasets.}
%\anurag{
We use Cityscapes~\cite{cordts2016cityscapes} and BDD100k~\cite{yu2018bdd100k} datasets for coarse annotations, as well as  Synscapes~\cite{wrenninge2018synscapes} and GTA-5~\cite{Richter_2016_ECCV} datasets for synthetic annotation. Statistics for these datasets are shown in~\cref{tab:stat}. Only the Cityscapes dataset provides manually annotated coarse data. As BDD100k does not include coarse annotations, we divide the 7,000 training samples into 3,000 fine annotated samples and 4,000 coarse samples where the coarse annotations are simulated (see supplement Sec.1 and Fig.1 for details). We report results on the standard validation set on Cityscapes and BDD100k. %\yongqin{mention the test set which should be fine annotations} - done %following Cityscapes split which has around 2975 fine annotated samples. 
%}

\myparagraph{Cost of annotation} 
The cost for fine annotation of Cityscapes images, where almost all pixels are annotated is around 90 minutes~\cite{cordts2016cityscapes} per image including quality control. On the contrary, the cost of coarse annotations is just 7 minutes~\cite{cordts2016cityscapes} per image. Further for BDD100k, fine annotations cost is around 75 min compared to just 7 min for coarse annotation per image. We get this annotation cost for BDD100k by manually annotating 10 samples each for coarse and fine annotations via labelme~\cite{russell2008labelme}. %Please note that the coarse annotation doesn't have proper class boundaries and a lot of pixels are not annotated, mostly around the class boundaries. 
We consider the cost of synthetic data annotation to be free following~\cite{paul2020domain}, as it is generated from photorealistic rendering techniques. %There may
%be overhead cost in quality control of the generated images but we ignore such cost in our experiments. 

%\myparagraph{Trusting Cityscapes Coarse Label}
%\label{mypg:cl}
%The cityscapes dataset also provides coarse annotation for fine-train and val set for which fine annotation is available. These coarse annotations match with the fine annotations with 97\% accuracy \cite{cordts2016cityscapes}. This suggests that the manual coarse annotation are good and can be trusted. The coarse annotation lacks finer details that are usually expensive to annotate.    

\myparagraph{Implementation details}
We use DeepLab-v3+ with Imagenet-pretrained Xception-71 as the backbone. We use SGD optimizer and ``Poly'' learning rate scheduler with power=2.0. We also set momentum to 0.9 and weight decay rate to 0.0001. We employ a crop size of $760 \times 760$ and a batch size of 12. For each round of training we train for 100 epochs. %We perform cross-domain augmentation with a probability of 0.5 for each training sample. 
Further we perform 3 iterations of our self training framework. For evaluation, we use multiscale inferencing with scales \{0.5,1,2\}. We use the same hyperparameters for both BDD100k and Cityscapes.

\begin{figure*}[t]
\centering
\includegraphics[width=0.92\linewidth]{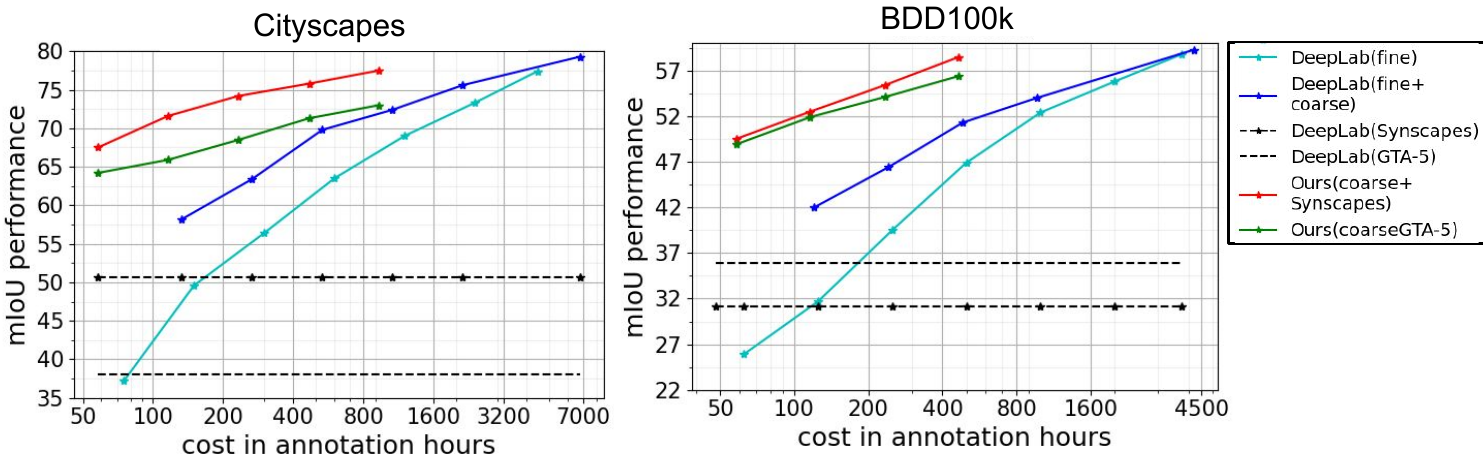}

\caption{Annotation cost vs performance. Baselines: 1.DeepLab(fine): the standard training strategy with fine data, 2.DeepLab~(fine+coarse): pretraining on coarse data following by fine-tuning on fine data, 3.DeepLab(Synscapes/GTA): standard training with synthetic Synscapes/GTA dataset, vs Ours~(coarse+syn(Synscapes/GTA-5)): our method trained on the combination of coarse and synthetic data. DeepLab-v3+ is abbreviated as DeepLab.  %The red plot presents our method which efficiently uses synthetic data along with coarse data to get better performance without any fine data. %\yongqin{update the legend: DeepLab-v3+(fine), DeepLab-v3+(fine+coarse), Ours(coarse+syn). For baselines, let's show the results until all the annotation hours. But for our approach, we only show the results until 1000 hours as result starts to decrease. } %\yongqin{It would be good to include the following results if we already have: ours(coarse + fine), ours(fine + syn), deeplab(coarse + syn), deeplab(fine + syn), or at least deeplab (coarse + syn)}
%\anurag{waiting for BDD results}
}
\label{fig:main_res}
\end{figure*}

{
 \setlength{\tabcolsep}{2pt}
 \renewcommand{\arraystretch}{1.2}
\begin{table*}

 \centering
 \resizebox{\linewidth}{!}{%
\begin{tabular}{l|l|c |c |c |c |c |c |c |c |c| c| c| c| c| c| c| c| c |c |c |c}
\hline
\multicolumn{22}{c}{Cityscapes} \\
\hline
Method & budget & \rotatebox{90}{road} & \rotatebox{90}{building} & \rotatebox{90}{vege.} & \rotatebox{90}{car} & \rotatebox{90}{sky} & \rotatebox{90}{sidewalk} & \rotatebox{90}{fence} & \rotatebox{90}{terrain} & \rotatebox{90}{wall} & \rotatebox{90}{pole} & \rotatebox{90}{tra. sign} & \rotatebox{90}{person} & \rotatebox{90}{bus} & \rotatebox{90}{truck} & \rotatebox{90}{bicycle} & \rotatebox{90}{train} & \rotatebox{90}{tra. light} & \rotatebox{90}{rider} & \rotatebox{90}{motor.} & mIoU \\
 %& & & &  & & & & light & sign & & & & & & & & & & cycle  &  \\
 
\hline
  %(coarse + synthetic) & & & & & & &&&&&&&&&&&&&& \\
 %\hline
 
  %ours~(coarse) & 933 & 96.4 & 75.3 & 88.8 & 50.7 & 53.4 & 52.4 & 59.6 & 69.4 & 87.8 & 46.9 & 92.6 & 70.0 & 54.3 & 90.0 & 79.5 & 83.2 & 78.0 & 56.9 & 67.7 & 71.2 \\
% \hline
 %\hline
 DeepLab-v3+~(fine) & 4462 & \textbf{98.0} & \textbf{92.2} & \textbf{92.2} & \textbf{94.9} & \textbf{94.8} & \textbf{84.5} & 58.4 & \textbf{62.7} & 56.8 & \textbf{61.2} & \textbf{76.1} & 80.4 & 88.7 & 81.9 & \textbf{75.2} & 80.6 & 65.8 & 60.6 & 62.4 & 77.4 \\
 
 Ours~(coarse + syn)  & \textbf{933} & 97.2 & 91.8 & 90.8 & 94.0 & 94.3 & 78.2 & \textbf{60.7} & 55.8 & \textbf{58.8} & 61.1 & 75.5 & \textbf{80.5} & \textbf{92.0} & \textbf{82.1} & 74.5 & \textbf{85.6} & \textbf{69.8} & \textbf{64.7} & \textbf{65.5} & \textbf{77.5} \\

%\hline
\hline
Pretrain (coarse) & 6795 & 98.3 & 92.7 & 92.5 & 95.3 & 95.1 & 86.4 & 65.2 & 61.4 & 54.2 & 65.9 & 78.9 & 81.9 & 90.7 & \textbf{86.3} & 77.1 & 83.8 & 69.4 & 64.8 & 67.5 & 79.3 \\
%pretraining & & & & & & &&&&&&&&&&&&&& \\
%\hline

Pretrain (ours) & 6795 & \textbf{98.4} & \textbf{93.5} & \textbf{93.2} & \textbf{96.0} & \textbf{95.5} & \textbf{87.3} & \textbf{66.3} & \textbf{65.1} & \textbf{54.7} & \textbf{71.2} & \textbf{82.5} & \textbf{85.5} & \textbf{92.6} & 84.5 & \textbf{80.8} & \textbf{88.0} & \textbf{75.3} & \textbf{70.7} & \textbf{72.7} & \textbf{81.8} \\
%pretraining & & & & & & &&&&&&&&&&&&&& \\
\hline
\hline
\multicolumn{22}{c}{BDD100k} \\
\hline
Method & budget & \rotatebox{90}{road} & \rotatebox{90}{car} & \rotatebox{90}{sky} & \rotatebox{90}{pole} & \rotatebox{90}{vege.} & \rotatebox{90}{building} & \rotatebox{90}{tra. sign} & \rotatebox{90}{sidewalk} & \rotatebox{90}{tra-light} & \rotatebox{90}{terrain} & \rotatebox{90}{person} & \rotatebox{90}{truck} & \rotatebox{90}{fence} & \rotatebox{90}{bus} & \rotatebox{90}{wall} & \rotatebox{90}{bicycle} & \rotatebox{90}{rider} & \rotatebox{90}{motor.} & \rotatebox{90}{train} & mIoU \\
\hline
 DeepLab-v3+~(fine) & \textbf{3750} & \textbf{94.6} & \textbf{90.2} & \textbf{95.2} & \textbf{48.3} & \textbf{85.7} & \textbf{85.4} & 49.2 & \textbf{62.5} & 48.3 & 44.3 & 64.3 & \textbf{57.5} & 45.7 & \textbf{75.5} & 27.1 & 47.8 & 46.3 & 48.8 & 0 & \textbf{58.8}
\\
 
 Ours~(coarse + syn) & 466 & 93.0 & 87.7 & 91.0 & 40.2 & 82.9 & 82.8 & \textbf{49.5} & 60.5 & \textbf{50.8} & \textbf{44.5} & \textbf{65.7} & 51.4 & \textbf{49.3} & 74.8 & \textbf{31.4} & \textbf{50.0} & \textbf{56.5} & \textbf{49.7} & 0 & 58.5
 \\

%\hline
\hline
Pretrain &  4216 & 94.7 & 89.7 & 95.2 & 46.5 & 85.7 & 84.9 & 48.2 & 63.7 & 47.1 & 46.9 & 61.2 & 56.2 & 49.6 & 77.9 & 37.8 & 46.9 & 43.7 & 50.5 & 0 & 59.3
 \\
%pretraining & & & & & & &&&&&&&&&&&&&& \\
%\hline

Pretrain (ours) &  \textbf{4216} & \textbf{95.3} & \textbf{91.2} & \textbf{95.6} & \textbf{57.1} & \textbf{87.1} & \textbf{86.8} & \textbf{58.9} & \textbf{67.7} & \textbf{59.7} & \textbf{48.3} & \textbf{69.9} & \textbf{58.3} & \textbf{51.9} & \textbf{82.9} & \textbf{35.3} & \textbf{59.3} & \textbf{55.4} & \textbf{55.9} & \textbf{0.4} & \textbf{64.1}
 \\
\hline
\end{tabular}
}
\caption{%\yang{Do we use Deeplabv3+ ?} 
Comparing with the best results of DeepLab-v3+. We report per-class IoU as well as mean IoU~(mIoU).
The class names are sorted in decreasing order of the class-wise image distribution. We show results for Cityscapes (top block) and BDD100k (bottom block). For each, we have two comparisons - 1) our results without using fine data vs DeepLab-v3+ uses all available fine data; our method performs better on tail classes with overall comparable performance at one-fifth (Cityscapes) and one-eighth (BDD100k) budget. 2) Pretraining with our framework vs pretraining with coarse data. Our model serves as a better pretraining method than directly coarse data pretraining.} %\yongqin{Do you have also sort the classes in BDD100k?}-- yes we sort the classes}%\anurag{waiting for BDD results}}
\label{tab:class}
\end{table*}
}

\subsection{Comparison with baselines}

In this section, we first introduce our baselines followed by showing the performance vs annotation cost trade-off and comparing with the best results of DeepLab-v3+. %We perform experiments using different combinations of real and synthetic data. Specifically, we show results for our framework, with coarse data from Cityscapes and BDD100k with synthetic data from Synscapes and GTA-5 individually.

% We present our maiwn results in two parts :
% \begin{itemize}[noitemsep,topsep=2pt,itemsep=1pt,leftmargin=8pt]
%     \item Low budget regime (see fig.  \ref{fig:main_res}) : We compare our framework's performance with standard ``fine training'' and ``fine with coarse pretraining'' in low annotation budgets.
%     \item Full budget comparison (see table \ref{tab:class}) : We compare our model performance with full fine supervision. We also present pretraining performance comparision between coarse coarse pretraining and pretraining with our framework.  
% \end{itemize}

\myparagraph{Baselines.} We take DeepLab-v3+~\cite{chen2018encoder} as the segmentation network and first compare with two popular supervision schemes adopted by most of existing works~\cite{chen2018encoder,zhang2021dcnas,yuan2021segmentation,yin2020disentangled}. The first baseline is trained with only fine data, denoted as DeepLab-v3+~(fine). The second baseline is trained on the combination of fine data and coarse data, denoted as DeepLab-v3+~(fine+coarse). Specifically, we first pretrain the network on the coarse data followed by fine-tuning it on fine data. Second, we also compare with another two intuitive baselines, DeepLab-v3+(Synscapes) and DeepLab-v3+(GTA) where model is trained with synthetic data. %For a fair comparison, our method is also with DeepLab-v3+, but is trained on the combination of coarse and synthetic data, denoted as Ours~(coarse + syn). 

\myparagraph{Performance vs annotation cost trade-off.} In this experiment, we compare with the baselines under different annotation costs on Cityscapes and BDD100k. For Cityscapes fine data, we draw 50, 100, 200, 400, 800, 1600 and 2975 images from the training set, taking 75, 150, 300, 600, 1200, 2400 and 4462 hours respectively to annotate.  Similarly, for coarse data, we draw 500, 1000, 2000, 4000 and 8000 images from the training set, amounted to 58, 116, 233, 467, and 933 annotation hours respectively. Finally, for synthetic data, we randomly draw 500, 1000, 2000, 4000 and 8000 images where we assume the annotation is free. %\yongqin{need to have to arguments on the zero cost here.--> gave reference to experiment section where we mention the zero cost}
Note that all those training examples for Cityscapes are sampled incrementally based on our proposed model-based sampling strategy described in \cref{subsec:select}. For BDD100k, we follow the same sampling scheme as Cityscapes.
%Following Cityscapes datasplits, for BDD100k we randomly sample 50, 100, 200, 400, 800, 1600 and 3000 fine images with annotation cost of 62.5, 125, 250, 500, 1000, 2000 and 3750 hours respectively. Similarly for coarse data from BDD100k, we sample 500, 1000, 2000 and 4000 images with annotation cost of 58, 116, 233 and 466 hours respectively. We use same synthetic datasplits for BDD100k as used for experiments with Cityscapes dataset.  \yongqin{We do not need to repeat numbers on BDD as they are exactly the same as Cityscapes! You do not need to mention 19998 as well. Too many integers numbers are just distracting and meaningless }

\cref{fig:main_res} shows the results on Cityscapes (left) and BDD100k (right) datasets. We have the following observations. In general, our method achieves the best annotation cost vs performance trade-off. 
The method ranking follows Ours(coarse +syn(Synscapes/GTA-5)) $>$ DeepLab-v3+~(fine+coarse)  $>$ DeepLab-v3+~(fine) $>$ DeepLab-v3+(synthetic) where synthetic can be GTA-5 or Synscapes. 
While using coarse data for pretraining, i.e., DeepLab-v3+~(fine+coarse), indeed improves fine data alone, i.e., DeepLab-v3+~(fine), our proposed coarse-to-fine framework achieves a significant larger performance boost.  
%consistently outperforming the competitors under different annotation costs. 
On Cityscapes, our method  with Synscapes obtains an impressive mIoU of $77.5\%$ with 933 annotation hours, which amount to only one-fifth of the annotation cost~(4462 hours) of using full fine data~(mIoU is $77.4\%$). This is encouraging because no prior works report such competitive results without using any fine data. 
%Similarly for BDD100k with Synscapes with just about one-eighth of fine budget (3750 vs 466 hrs), our method performs comparable to DeepLab-v3+ baseline  ($58.5\%$ mIoU of ours vs $58.8\%$ mIoU for DeepLab-v3+~(fine)).
%Moreover, we observe that our method using synthetic data from Synscapes performs better than with GTA-5, which is expected because Synscapes is more similar to Cityscapes. Nevertheless, our method is able to improve the baselines with both synthetic datasets, indicating that our method is not limited to similar synthetic data. Similar conclusions can be observed on BDD100k dataset.

%is able to outperform the baselines with different synthetic data i.e., Synscapes and GTA-5. This suggests that our method is not limited to a particular synthetic data, as our method performs well with using GTA-5 data, which is very different from Synscapes.

%When we use GTA-5 as synthetic data, our method still outperfoms the baseline with both Cityscapes and BDD100k data. Specifically, for BDD100k, it outperforms the baseline by a gap of 9.4\% mIoU with annotation cost of 466 hrs compared to 500 hrs for the baseline.

In addition, we find that the performance gap is larger under a small annotation budget e.g., for Cityscapes experiment, ours achieves $67.5\%$ mIoU with 58 annotation hours vs DeepLab-v3+~(fine)'s $37.2\%$ mIoU  with 75 annotation hours. Similarly for BDD100k, ours achieve $49.9\%$ mIoU with 58 annotation hrs vs $25.9\%$ mIoU with 62.5 annotation hrs. This implies that annotating diverse coarse examples is much more important than carefully labelling every pixel when the annotation budget is small, shedding light on annotating large-scale urban scene segmentation datasets. More importantly, these results empirically confirm the effectiveness of our proposed coarse to fine framework. 
\myparagraph{Comparing with the best results of DeepLab-v3+.} We present the best results achieved by DeepLab-v3+ in ~\cref{tab:class} for both Cityscapes and BDD100k datasets. For Cityscapes, 
%We also compare our framework's performance with full budget baseline's performance in \cref{tab:class}.
Compared to DeepLab-v3+~(fine) trained on full training set~(~\cref{tab:class}), Ours~(coarse+syn) is able to achieve a competitive result with only one-fifth of its cost~(933 vs 4462 hours). For BDD100k, we achieve comparable performance with only one-eighth of cost wrt full fine training~(466 vs 3750 hours). Interestingly, our method tends to perform better on the tail classes with a lower number of pixel instances. For example, our method substantially outperforms DeepLab-v3+~(fine) on train, traffic light, rider and motor for Cityscapes, and wall, bicycle, rider and motorcycle for BDD100k. This is attributed to the improved diversity in training samples from our cross-domain data augmentation.
%and our model-based balance sampling. %We observe that our model performs better than full fine supervision in almost one-fifth (933 vs 4462) on the full annotation budget (77.5 vs 77.4). 
%This is motivating as we didn't use any fine data and achieved better performance with just using coarse and synthetic data which earlier were used only for pretraining purposes.
%To further improve the results, we initialize the DeepLab-v3+ with a pretrained model followed by fine-tuning it on the full fine data. Here, 
We also compare two pretraining strategies: (1) the model trained on coarse data following the conventional way~\cite{chen2018encoder}, (2) the model trained on coarse and synthetic data using our framework.  %Specifically, we initialize DeepLab-v3+ with the model pretrained on coarse data in the conventional way~\cite{chen2018encoder}. 
%We compare standard coarse pretraining with pretraining from our framework. 
As shown in \cref{tab:class},  pretraining with our framework leads to a significant improvement of 2.5 mIoU for Cityscapes and 4.8 mIoU for BDD100k datasets. These results imply that our method could serve as a promising pretraining strategy for the best performance.  

\begin{figure*}[t]
\centering
\includegraphics[width=0.24\linewidth,trim=17 0 30 0,clip]{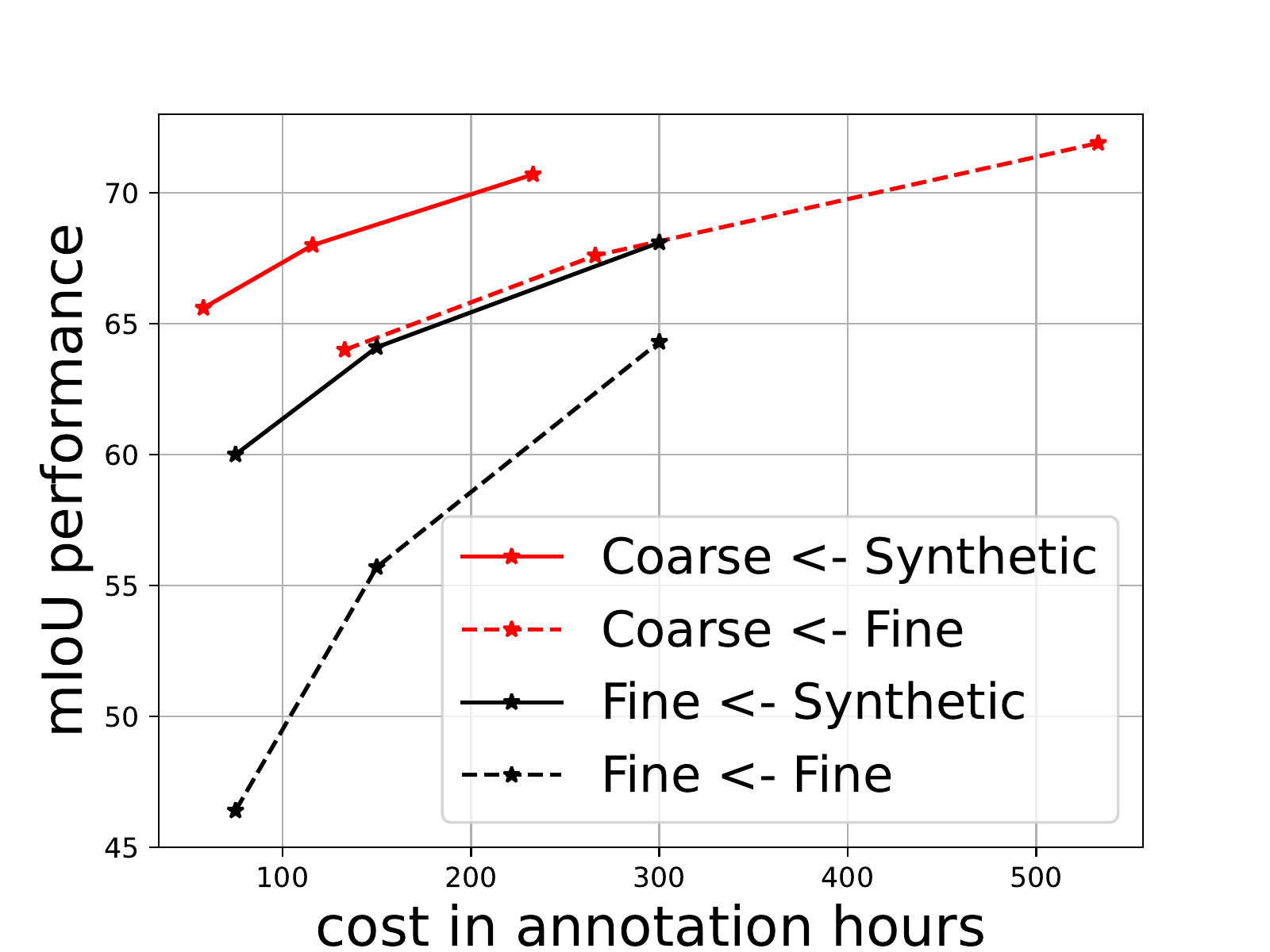}
\includegraphics[width=0.24\linewidth, trim=17 0 30 0,clip]{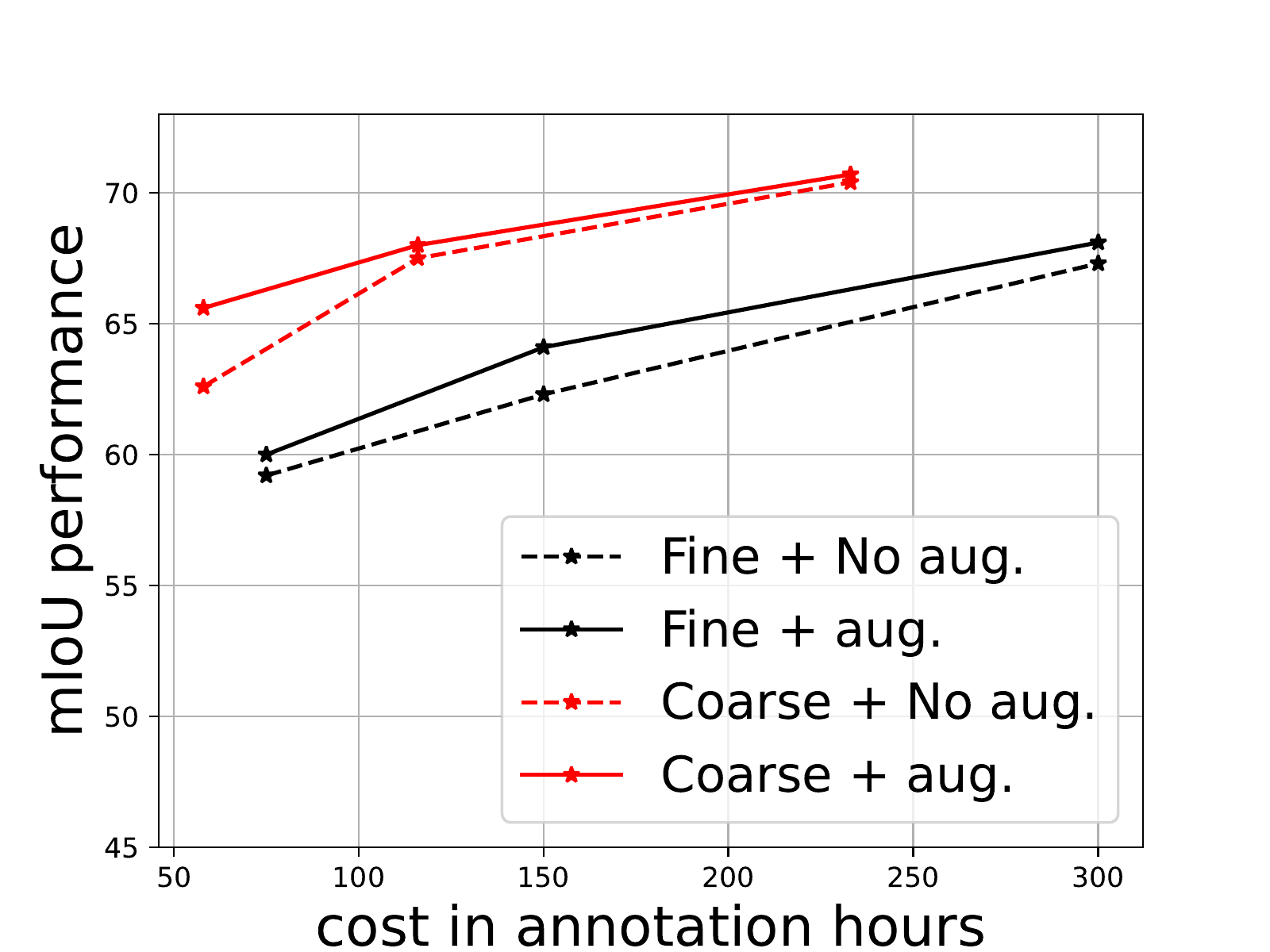}
\includegraphics[width=0.24\linewidth, trim=17 0 30 0,clip]{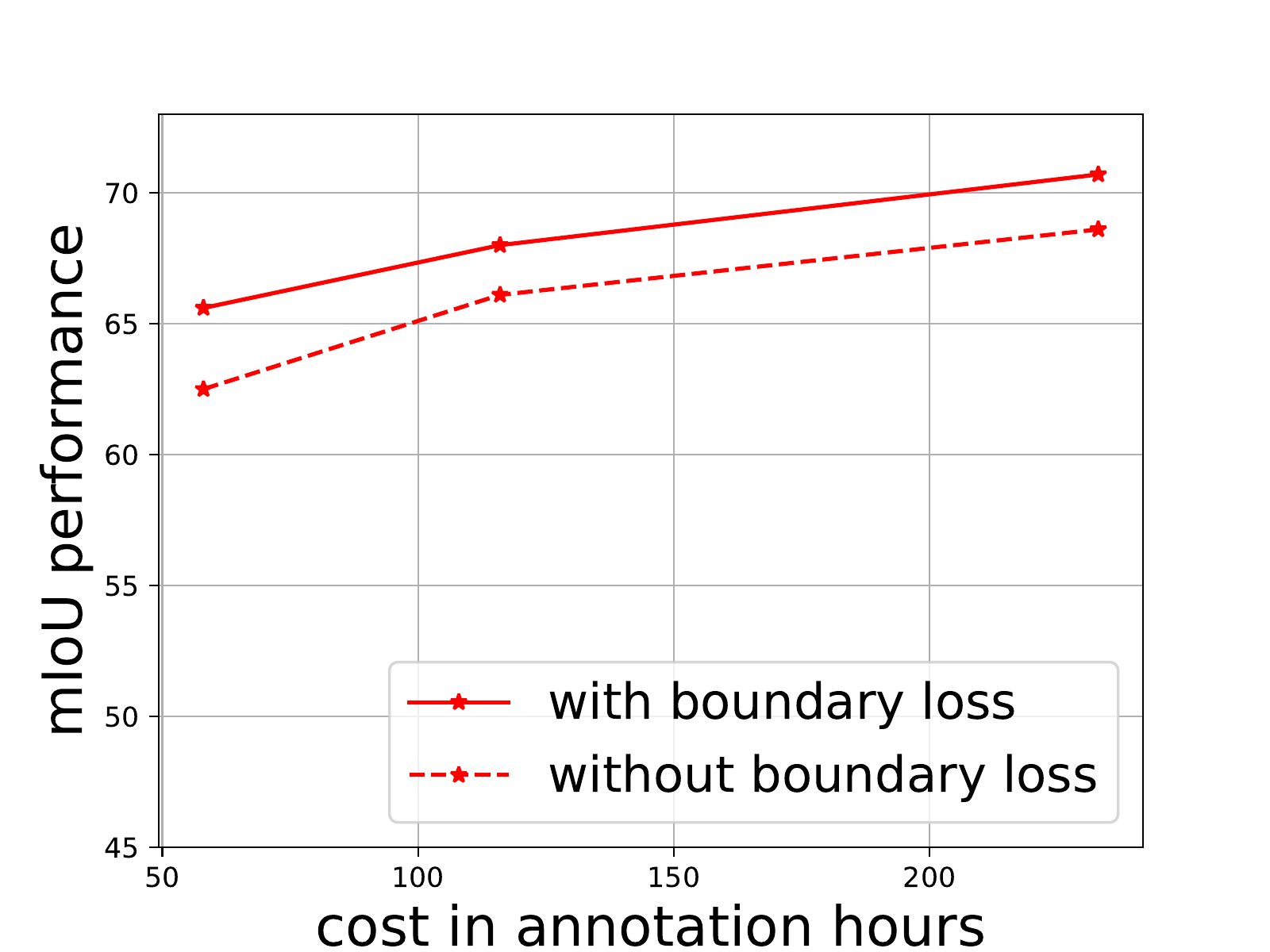}
\includegraphics[width=0.24\linewidth, trim=23 23 20 0,clip]{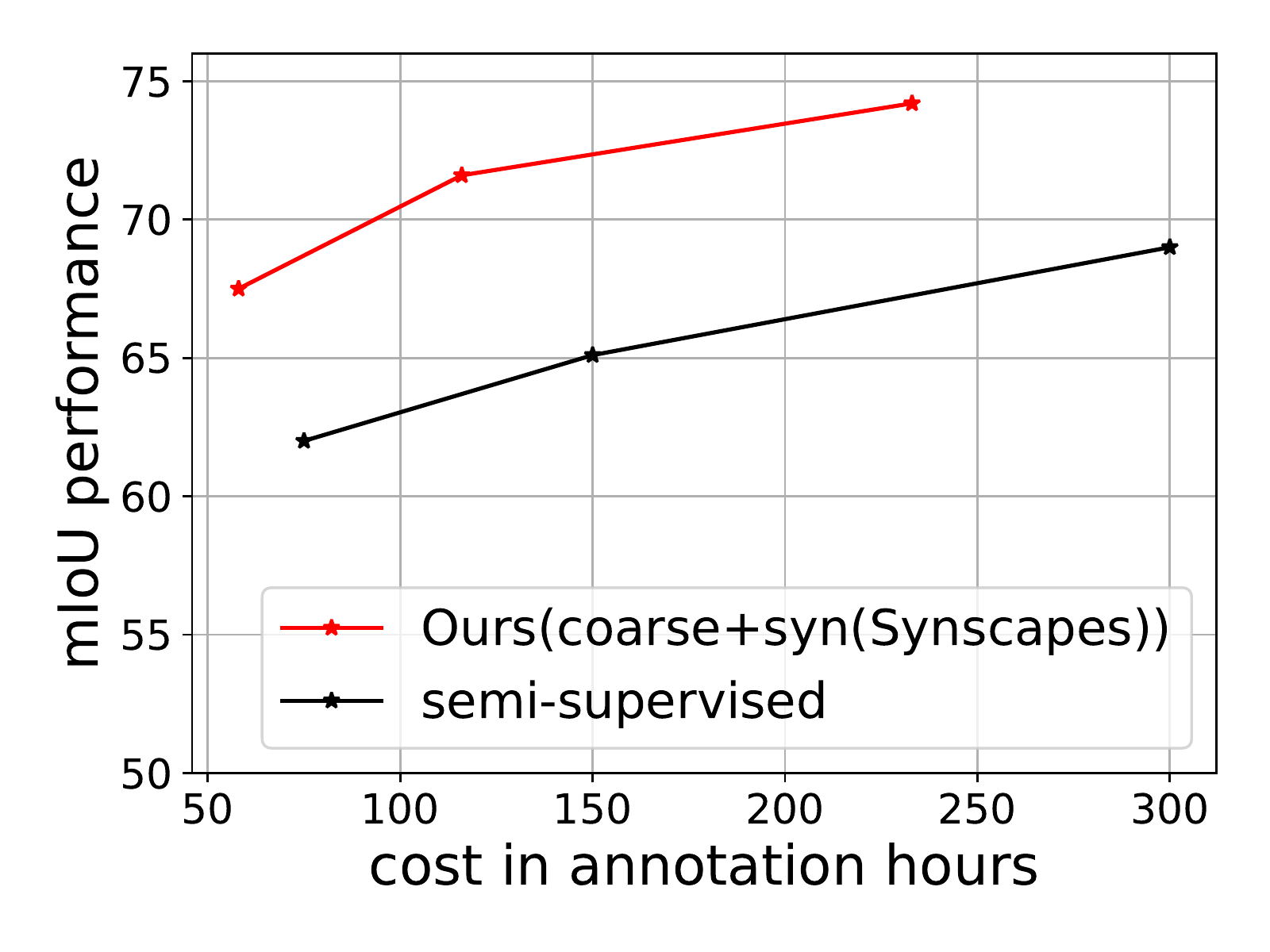}
\caption{Left to Right: 1) Synthetic vs Fine for cross domain augmentation. 2) Importance of cross domain augmentation. 3) Importance of boundary loss. 4) Comparison with a semi-supervised learning baseline~\cite{chen2020naive} on Cityscapes.  %The red solid line (ours) is common for all three plots. We ablate wrt to this plot.
}
\label{fig:ablation1}
\end{figure*}
%\begin{figure}
%\centering

%\includegraphics[width=0.7\linewidth, trim=10 0 10 0,clip]{images/costvsperf_abl10.pdf}
%\caption{Comparison with a semi-supervised learning baseline~\cite{chen2020naive} on Cityscapes. %\yongqin{remove with pspnet in the x-axis label. also, cite the semi-supervised method.} %The red solid line (ours) is common for all three plots. We ablate wrt to this plot.
%}
%\label{fig:ablation2}
%\end{figure}

\subsection{Ablation study on model components and hyperparameters}
In this section, we conduct the ablation studies on cross-domain augmentation, boundary loss, impact of self-training iterations and comparison with a semi-supervised learning method. The ablation experiments are conducted on Cityscapes with Synscapes being the synthetic dataset.

% \begin{figure}[t]
% \centering
% \includegraphics[width=\linewidth]{latex/images/costvsperf_abl3.pdf}
% \caption{Importance of DACS based augmentation. The black solid and dotted plot represents using fine data with and without DACS based augmentation respectively. Similarly, red solid and dotted plot represents using coarse data with and without DACS based augmentations. For both cases, using DACS based augmentation helps in improving the performance. }
% \label{fig:data_aug}
% \end{figure}

\myparagraph{Effect of cross domain augmentation.} In our method, we perform the cross-domain augmentation by copy-pasting synthetic objects onto coarse images i.e., Coarse $\leftarrow$ Synthetic. In this study, we ablate other augmentation choices given two domains of data i.e.,  Coarse $\leftarrow$ Fine, Fine $\leftarrow$ Synthetic,  Fine $\leftarrow$ Fine. As shown in \cref{fig:ablation1} (left), applying augmentation from Synthetic to Coarse data works the best. We also perform ablation to study the importance of cross-domain augmentation for our self-training framework. We present ablation results for both fine and coarse data in \cref{fig:ablation1} (mid). We observe improvement in performance by using cross-domain augmentation for both fine and coarse data. In particular, the improvement for the lowest budget case (i.e., 58 hours) using coarse data is siginificantly improved (by a mIoU of 3\%) by mixing data from Synscapes, where it provides sufficient samples with important details not present in the original data with coarse annotation.
%We present ablation results for both fine and coarse data in . 
%We use augmentation from Synscapes dataset as discussed in \cref{pg:cda}. For fine data we observe a performance improvement of around 0.8, 1.8 and 0.8 mIoU at the budgets of 75, 150 and 300 annotation hours respectively, which shows the effectiveness of the augmentation. Similarly, we observe improvement of 3.0, 0.5, 0.3 mIoU for coarse data at the annotation budgets of 58, 116 and 233 hours. In particular, the improvement for the lowest budget case (i.e., 58 hours) using coarse data is siginificantly improved by mixing data from Synscapes, where it provides sufficient samples with important details not present in the original data with coarse annotation. To conclude, DACS data augmentation is quite helpful to improve our coarse-to-fine system, which trains a segmentation using coarse annotation.

% \begin{figure}[t]
% \centering
% \includegraphics[width=\linewidth]{latex/images/costvsperf_abl5.pdf}
% \caption{Importance of boundary loss. The red (solid) plot show model performance with boundary loss and red (dotted) plot shows performance without boundary loss.}
% \label{fig:bound_loss}
% \end{figure}

\myparagraph{Effect of boundary loss.}
We present the ablation of the boundary loss in \cref{fig:ablation1}~(mid). Even though synthetic data provides necessary boundary information, we can observe that training a network with only cross-entropy loss performs worse than additionally applying the boundary loss. Specifically, we observe performance gains of 3.1, 1.9 and 2.1 mIoU compared with the non-boundary version for annotation budgets of 58, 116 and 233 hrs, which corresponds to using 500, 1000 and 2000 images with coarse annotation.  

%{
 %\setlength{\tabcolsep}{4pt}
 %\renewcommand{\arraystretch}{1.2}
%\begin{table}

 %\centering
 %\resizebox{0.6\linewidth}{!}{%
%\begin{tabular}{l|c c c c c c}
% \textbf{Coarse Samples} & 500 & 1K & 2K & 4K & 8K & All \\ 
%\hline
%Iteration 0 & 65.6  & 68.0 & 70.7 & 73.4 & 74.5 & 74.7 \\
%Iteration 1 & 67.1  & \textbf{71.6} & \textbf{74.2} & \textbf{75.8} & 76.6 & \textbf{77.0} \\
%Iteration 2 & \textbf{67.5} & 70.4 & 73.4 & 74.4 & \textbf{77.5} & 76.8 \\
%\hline
%\end{tabular}
%}
%\caption{Self-training iterations using a mixture of coarse annotated data and synthetic data. Performance in %different self-training iterations of our  framework. Our framework achieves optimal performance in iteration 1 %for most experiments itself and we do not need to perform more iterations.}
%\label{tab:iteration}
%\end{table}
%}

{
 \setlength{\tabcolsep}{4pt}
 \renewcommand{\arraystretch}{1.1}
\begin{table}
 \centering
 \resizebox{0.8\linewidth}{!}{%
\begin{tabular}{l|c c c c c }
 
 \textbf{Coarse Samples} & 500 & 1K & 2K & 4K & 8K  \\ 
\hline
Iteration 0 & 65.6  & 68.0 & 70.7 & 73.4 & 74.5 \\
Iteration 1 & 67.1  & \textbf{71.6} & \textbf{74.2} & \textbf{75.8} & 76.6  \\
Iteration 2 & \textbf{67.5} & 70.4 & 73.4 & 74.4 & \textbf{77.5}  \\
\hline
\end{tabular}
}

\caption{Self-training iterations using a mixture of coarse annotated data and synthetic data. Performance in different self-training iterations of our  framework. Our framework achieves optimal performance in iteration 1 for most experiments itself and we do not need to perform more iterations.}
\label{tab:iteration}
\end{table}
}

{
 \setlength{\tabcolsep}{4pt}
 \renewcommand{\arraystretch}{1.1}
\begin{table}

 \centering
 \resizebox{0.7\linewidth}{!}{%
\begin{tabular}{l|c c c c}
 \textbf{Coarse Samples} & 1K & 2K & 4K & 8K  \\ 
\hline
Model-based & 57.4  & \textbf{62.3} & \textbf{66.4} & \textbf{68.4} \\
uniform sampling & 57.4  & 59.7 & 64.5 & 66.5 \\
\hline
\end{tabular}
}

\caption{ Model-based class balanced sampling vs random sampling. Our model based iterative sampling generates diverse training samples compared to uniform sampling.}
\label{tab:sampling}
\end{table}
}

\myparagraph{Number of self-training iterations.} The last two stages of our coarse-to-fine self-training framework can be trained iteratively. In \cref{tab:iteration},  we present the performance improvement at different iterations with 500, 1000, 2000, 4000 and 8000 coarse samples. Iteration 0 corresponds to network pre-training stage, as explained in \cref{alg:self_tr}. After iteration 0, we generate pseudo labels for the ignored regions in the coarse annotated images and create new GT for the next iteration. There are clear improvements from iteration 0 to 1, as the model touches newly annotated pixels which are ignored in the iteration 0. At iteration 2, the performance may slightly decrease in some cases due to the inevitable errors in pseudo labels. %These results demonstrate that performing multiple self-training iterations is able to gradually improve the performance. 
%Further, we also obtain 77.5 mIoU with just 933 annotation hours, which is better than 77.4 mIoU using all fine annotations with the budget of 4462 annotation hours. To conclude, we believe our coarse-to-fine framework is effective to train a semantic segmentation model with low-cost annotations.

%{
% \setlength{\tabcolsep}{8pt}
% \renewcommand{\arraystretch}{1.2}
%\begin{table}

 %\centering
 %\resizebox{0.6\linewidth}{!}{%
%\begin{tabular}{l|c c c c}
 %\textbf{Coarse Samples} & 1K & 2K & 4K & 8K  \\ 
%\hline
%Model-based sampling & 57.4  & \textbf{62.3} & \textbf{66.4} & \textbf{68.4} \\
%random & 57.4  & 59.7 & 64.5 & 66.5 \\
%\end{tabular}
%}
%\caption{Model-based class balanced sampling vs random sampling. No synthetic data with fine annotation is used.}
%\label{tab:cb}
%\end{table}
%}

\myparagraph{Which examples should be labeled?} Coarse annotation is not expensive to acquire, but it is interesting to know which examples to label earlier.
We gradually expand the training examples from 1000 to 8000. At every step, we sample new coarse examples to train a model as discussed in \cref{subsec:select}.
%, we first uniformly sample 1000 coarse examples and train our model to obtain the class distribution. We use the class distribution to incrementally add new samples. In every step, we apply the model trained on the previous step to select examples for our method. 
We compare our sampling strategy with uniform sampling in~\cref{tab:iteration}. Apparently, we can see the effectiveness of the model-based sampling and better results can be obtained.

%we observe that our sampling method is efficient and improve performance compared to random sampling without adding overhead classification cost.   

%\myparagraph{Using our trained model for pretraining.} \\
%\label{pg:pretrain}
%We can also use the model trained from our work for pretraining. With coarse pretraining we obtain a performance of 79.3 when use all of the fine annotation data. With using our trained model for pretraining we achieve a high gain with performance of 81.8 mIoU on val dataset with same fine annotations. \anurag{not sure if we should keep this, but the number seemed quite high for just pretraining with coarse + synscapes with deeplabv3+ backbone. We do not have results with normal coarse + synscapes pretraining and it will take time to run it. Not possible before deadline. \anurag{annotation cost column, another col for fine, coarse, performance, all classes --- 2 columns}} 

% \begin{figure}[t]
% \centering
% \includegraphics[width=0.8\linewidth]{latex/images/costvsperf_abl7.pdf}
% \caption{Ablation study with different segmentation network (PSPNet~\cite{zhao2017pyramid}). Ours refer to our full model, Baseline refers to model trained on coarse + synthetic data, ``Baseline + boundary'' adds aditional boundary loss and ``baseline + DACS'' adds augmentation on top of baseline. }
% \label{fig:pspnet}
% \end{figure}

\begin{figure*}[t]
\centering
\includegraphics[width=0.9\linewidth]{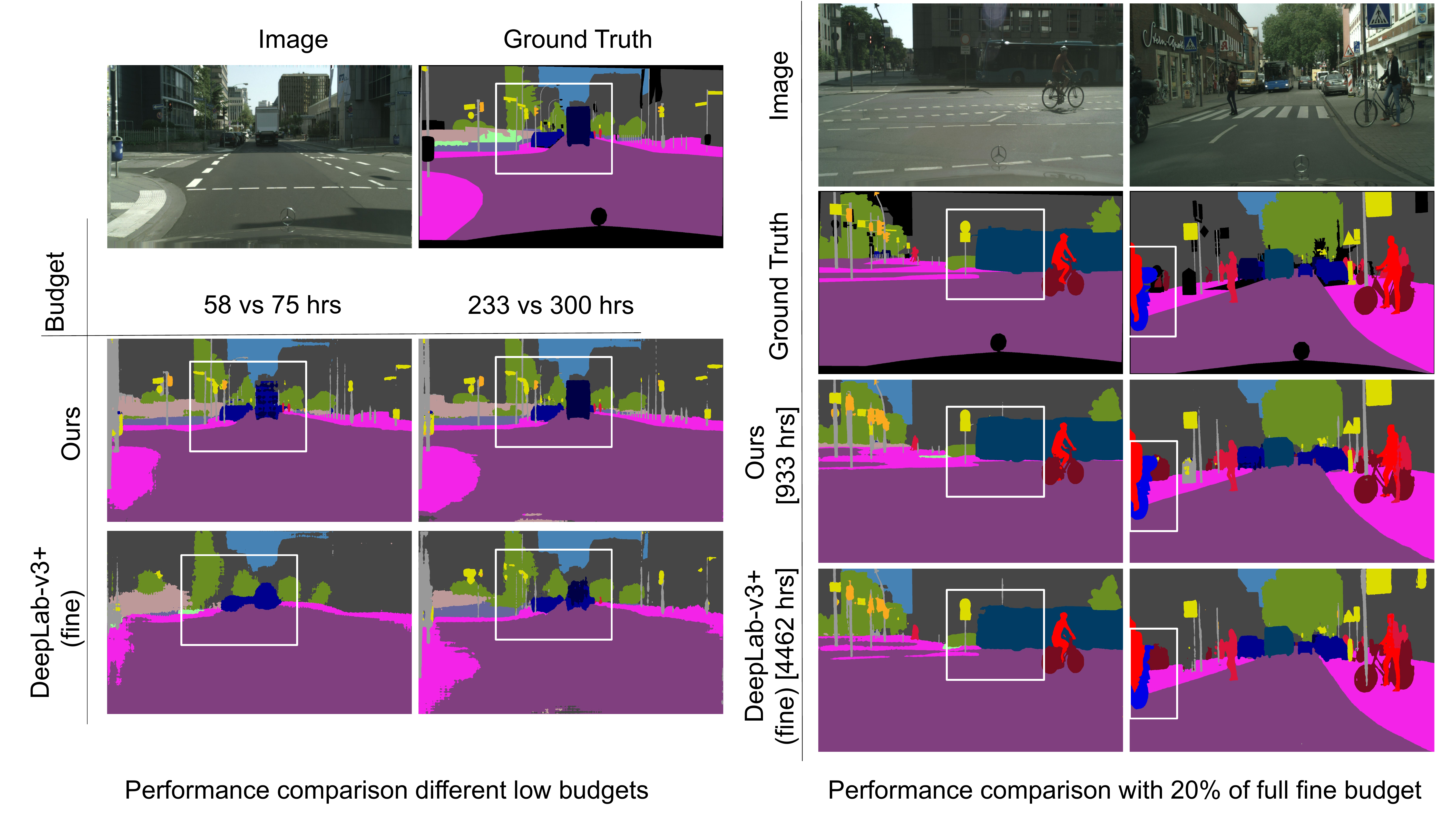}
%https://docs.google.com/drawings/d/1gk9qawecy7-TOeKMmVUwkn0WN9Utze4OB-WmplQST6Y/edit?usp=sharing
\caption{ %\yang{Try to highlight some regions that our approach is better, for example poles}
Qualitative comparison of ours vs DeepLab-v3+(fine) as baseline on Cityscapes dataset. Left: Qualitative performance for an image on different annotation budgets i.e. 58(ours) vs 75(baseline) and 233(ours) vs 300(baseline) annotation hours. Right: Qualitative comparison with 20\% of full fine budget (933 vs 4462 annotation hours). Even with 5 $\times$ cheaper budget, our coarse-to-fine framework performs comparable with full fine annotation budget. Further, on some tails end classes, it even performs better(e.g., motorbike, bus, as highlighted). Region of interests is present in white bounding boxes.  
}
\label{fig:qualitative}
\end{figure*}

\myparagraph{Comparison with semi-supervised learning.} We also compare our learning framework with a semi-supervised learning method~\cite{chen2020naive}  in \cref{fig:ablation1}(right). We adapt the codes provided by \cite{chen2020naive} for this comparison. For this experiment, we use the same coarse data samples for a given budget as our other experiments. We treat the coarse data samples as unlabelled data by not using its annotation mask. Similary, We use same the fine data samples as used in our baselines for a given budget.%We use coarse data without any labels as the unlabelled data. We use fine annotation samples for learning, and generate pseudo labels following~\cite{chen2020naive} for unlabelled . Next, we iteratively train the model with both pseudo labels and available fine annotation. 
Our framework achieves 67.5\% in a budget of 58 annotation hours vs 62.0\% by semi-supervised approach with 75 hrs annotation budget (see ~\ref{fig:ablation1}(right)). Similarly, our framework with a budget of 116 and 233 annotation hours outperforms the semi-supervised approach by a gap of 2.4\% and 1.4\% mIoU with budgets 150 and 300 hours respectively. This shows the importance of coarse annotation, which along with synthetic data can be efficiently used to obtain competitive performance. 

\subsection{Qualitative results}
We visualize the qualitative comparison of the baseline with our framework in \cref{fig:qualitative}. In the left,  we provide a comparison with the baseline trained with the fine annotated data at the  budgets of 75 and 300 hours, which is slightly more than our versions at 58 and 233 hours. We observe that the performance of small objects and the region with rich details are improved. For instance, we highlight the truck and the traffic-light are predicted correctly from our model with 233 hours, while the model using fine annotation cannot recognize them well. Furthermore, our model with 58 hours can still recognize the majority of those objects, while the comparing method with 75 hours directly ignore those objects even though its training images have labels for every pixel.
%, which is incorrectly predicted by deeplab-v3+ for both budgets. For 75 annotation hours budget, our framework is able to predict the truck sparsely, but for more budget i.e. 300 annotation hours, it does correct prediction for whole truck. 
Also, baseline fails to predict traffic-light for both budgets, while our method succeeds. In the right, we provide a comparison with full fine annotation budget. Our framework with only one-fifth of full budget (933 vs 4462) is able to perform equally well with baseline as can be observed qualitatively. Moreover, the qualitative results also confirm that our framework performs better on tail classes (see \cref{tab:class}). For instance bus and motorcycle (right, column1 and column 2 respectively) compared to baseline. 
%Our framework does better prediction for tail-end classes bus and motorcycle (right, column1 and column 2 respectively) compared to DeepLab-v3+. 

%\anurag{model figure - sample figure}

%teaser - figure (coarse, fine)
%two approaches - std fine - money
%ours - coarse + synthetic less money

%TODO : cost table - general statistics - pixel percent annotation, 
\section{Conclusion}
\label{sec:concl}

In this work, we argue that coarsely annotated data has been largely ignored  as a primary training source. Therefore, we propose a new supervision scheme based on coarse data and synthetic data, significantly reducing annotation time. We develop a strong baseline that efficiently learns from coarse and synthetic data by combing self-training, a boundary loss and cross-domain data augmentation. We conduct extensive experiments to evaluate our method on Cityscapes and BDD100k datasets with two different synthetic datasets i.e., Synscapes and GTA-5. The experimental results show that our method achieves the best performance vs. annotation cost trade-off when compared to the standard supervision schemes with fine annotated data. More importantly, our method achieves competitive performance for Cityscapes compared to the state of the art with only one-fifth of its annotation budget. We hope our method inspires more future works along this challenging but rewarding research direction.

%\myparagraph{Limitations.} One potential limitation is that considering the cost for synthetic data as free is probably not entirely true. But as synthetic data can be used for many tasks this might be still considered to be approximately correct. 

{\small
\bibliographystyle{ieee_fullname}
\bibliography{egbib}
}

\end{document}